\documentclass{bmvc2k}

\title{JOLT3D: Joint Learning of Talking Heads and 3DMM Parameters with Application to Lip-Sync}

\addauthor{Sungjoon Park}{fatrat@snu.ac.kr}{1}
\addauthor{Minsik Park}{mspark@hudson-ai.com}{1}
\addauthor{Haneol Lee}{haneol.lee@hudson-ai.com}{1}
\addauthor{Jaesub Yun}{jaesub.yun@hudson-ai.com}{1}
\addauthor{Donggeon Lee}{donggeon.lee@hudson-ai.com}{1}

\addinstitution{
 Hudson AI\\
 Seoul, KR
}

\runninghead{Park et al.}{JOLT3D}

\usepackage{amssymb}
\usepackage{hyperref}
\usepackage{booktabs}
\usepackage{url}
\usepackage{mathtools}
\usepackage{listings}
\usepackage{xcolor}
\usepackage{algorithm}
\usepackage{algpseudocode}

\newcommand{\red}[1]{\textcolor{red}{#1}}

\begin{document}

\maketitle

\begin{abstract}
In this work, we revisit the effectiveness of 3DMM for talking head synthesis by jointly learning a 3D face reconstruction model and a talking head synthesis model.
This enables us to obtain a FACS-based blendshape representation of facial expressions that is optimized for talking head synthesis.
This contrasts with previous methods that either fit 3DMM parameters to 2D landmarks or rely on pretrained face reconstruction models.
Not only does our approach increase the quality of the generated face, but it also allows us to take advantage of the blendshape representation to modify just the mouth region for the purpose of audio-based lip-sync.
To this end, we propose a novel lip-sync pipeline that, unlike previous methods, decouples the original chin contour from the lip-synced chin contour, and reduces flickering near the mouth.
Project page: \url{https://park-sungjoon.github.io/jolt3d/}
\end{abstract}

%-------------------------------------------------------------------------
\section{Introduction}
The 3D morphable face model (3DMM) \cite{egger20203d} is an important and widely used tool for modeling the human face.
With the advent of deep learning, the task of monocular 3D face reconstruction from in-the-wild face images using the 3DMM has been extensively studied \cite{tewari2017mofa,deng2019accurate,li2020learning,danvevcek2022emoca,chai2023hiface}.
This has motivated the use of 3DMMs for talking head synthesis (THS).
Although many efforts have been made in this direction, the idea has not gained popularity.
This is because existing approaches typically extract the 3DMM parameters either via landmark fitting \cite{doukas2021headgan,zhang2021flow,ye2024real3d} or through pretrained models \cite{yi2020audio,wu2021imitating,ren2021pirenderer,ye2022dynamic,xu2023high,zhang2023sadtalker,ji2024realtalk,zeng2023face,wang2025joygen}.
The former approach is inherently limited, as 2D landmarks cannot fully capture the 3D geometry of the face.
This limitation is exacerbated by the fact that off-the-shelf landmark detectors are often trained on noisy human-annotated labels.
The latter approach is constrained by the quality of pretrained 3D face reconstruction models—referred to as ReconNet in this work—which may not yield satisfactory results.
In particular, off-the-shelf ReconNets~\cite{deng2019accurate,feng2021learning,danvevcek2022emoca} are typically trained using photometric losses by utilizing the texture map of the Basel Face Model~\cite{paysan20093d,gerig2018morphable}.
This is suboptimal because these 3DMM parameters are not optimized for THS, and the texture map of the Basel Face Model cannot accurately capture the facial attributes.

Moreover, using a pretrained ReconNet limits the flexibility of adopting a different 3DMM than the one it was originally trained with.
This is because one must not only retrain the ReconNet, but also consider the fact that not all 3DMMs come with a texture map.
Popular ReconNets such as D3DFR~\cite{deng2019accurate}, DECA~\cite{feng2021learning}, and EMOCA~\cite{danvevcek2022emoca} are based on FLAME~\cite{li2020learning} or FaceWarehouse~\cite{cao2013facewarehouse}.
However, the expression parameters of FLAME are hard to interpret because they are generated by PCA.
Since there are many alternatives to the 3DMM for encoding the facial attributes, such as neural embeddings~\cite{drobyshev2022megaportraits,drobyshev2024emoportraits,xu2025vasa} or keypoints~\cite{siarohin2019first,wang2021one,guo2024liveportrait}, FLAME offers limited advantage over these alternatives.
The FaceWarehouse uses blendshapes based on Facial Action Coding System~\cite{ekman1978facial} (FACS), but does not have eyeball mesh, which makes gaze control difficult without additional modeling.

In this work, we propose to jointly train the ReconNet and the THS model.
To the best of our knowledge, this is the first effort towards training a ReconNet that is optimized for THS.
Instead of using FLAME or FaceWarehouse, we adopt the ICT-FaceKit \cite{li2020learning}, which has 55 blendshapes (compared to 46 in FaceWarehouse) and includes eyeball geometry.
We prefer ICT-FaceKit because the FACS-based blendshapes enable intuitive control of the face.
The eyeballs also allow gaze control without additional modeling~\cite{doukas2023free}.
Although ICT-FaceKit lacks a texture map, this is not a hindrance due to our training strategy.
We therefore free 3DMM-based THS from the limitations outlined above.

Thanks to the disentangled control of facial regions enabled by FACS-based blendshapes, it is possible to modify only the mouth-related blendshapes for lip-sync applications.
For this, we train a model to predict mouth blendshapes from audio and talking style using a diffusion model~\cite{sohl2015deep,song2020score,ho2020denoising}.
The face generated by using these mouth blendshapes can then be blended into the original video by a simple face blending network.
However, blending the generated face into the original video is complicated by the choice of the mouth mask.
Common approaches include masking the lower half of the face~\cite{prajwal2020lip}, using a fixed square region~\cite{zhang2023dinet}, or employing a mouth-shape-agnostic mask~\cite{li2024latentsync}.
We find these strategies unsatisfactory, and propose instead to inpaint just the mouth region after modifying the chin contour of the original face, see Fig.~\ref{fig:fig4} for an overview of the lip-sync pipeline.

The main contributions of this work are as follows:
\begin{itemize}
    \item We propose JOLT3D, a novel framework where ReconNet is optimized for THS.
    \item We achieve disentangled control over FACS-based blendshapes, including gaze.
    \item We propose a novel lip-sync pipeline that avoids the common mouth-mask artifacts.
\end{itemize}

\section{Related works}
\subsection{Audio-driven talking head synthesis}
Early works directly mapped audio to lip shapes by encoding and decoding the reference images and audio, constrained by a lip-sync loss~\cite{prajwal2020lip,park2022synctalkface,muaz2023sidgan,wang2023seeing}.
Other approaches predicted 3DMMs~\cite{zhang2021facial,zhang2021flow,zhang2023sadtalker,ji2024realtalk,ye2024real3d} or facial landmarks~\cite{zhong2023identity} from audio for THS.
Recent methods use diffusion models to predict various forms of facial representations from audio \cite{xu2025vasa,liu2024anitalker,xu2024hallo,jiang2024loopy,li2024latentsync} for THS.
In our work, 3DMM is used as the facial representation for THS, and diffusion model is used to predict the 3DMM parameters from audio and talking style.

\subsection{Talking head synthesis based on feature warping}
In this family of methods, faces are generated by decoding the warped features extracted from reference images.
Note that because convolution is translation-equivariant, smoothly varying warping fields tend to produce temporally consistent results.
Some methods constrain the warping field to affine transformations~\cite{zhang2021flow,zhang2023dinet}.
Implicit keypoints, which can be either 2D~\cite{siarohin2019first,ji2022eamm} or 3D~\cite{wang2021one,zhang2023sadtalker,guo2024liveportrait}, are also commonly used to guide the warping process. 
Other approaches rely on dense warping fields, which can be either 2D~\cite{doukas2021headgan,doukas2023free} or 3D~\cite{drobyshev2022megaportraits,drobyshev2024emoportraits,xu2025vasa}.
The neural embeddings used in MegaPortraits~\cite{drobyshev2022megaportraits} can capture detailed expressions~\cite{drobyshev2024emoportraits}, although it requires careful treatment of the latent vectors.
Finally, some methods propose warping the features implicitly~\cite{mallya2022implicit,ye2024real3d}.

In our work, we jointly train a ReconNet with a THS model that is based on dense 2D warping fields.
This turns out to be very effective, and we expect that similar compatibility will hold for THS methods based on 3D warping fields.
% In our work, we use dense 2D warping fields and demonstrate that learning 2D warping fields is highly compatible with the task of face reconstruction.
% Although it is not explored in this work, we expect similar compatibility to hold for the 3D warping field based methods.

\subsection{Talking head synthesis that uses 3DMMs}
While numerous studies use 3DMMs for THS, they typically either fit the 3DMM parameters to 2D facial landmarks~\cite{zhang2021flow,doukas2021headgan,ye2024real3d}, or use a pretrained ReconNet to extract the 3DMM parameters~\cite{yi2020audio,wu2021imitating,zhang2021facial,ren2021pirenderer,ye2022dynamic,xu2023high,zhang2023sadtalker,zeng2023face,sun2024uniavatar,wang2025joygen}.
In contrast, our approach jointly trains ReconNet with the talking head generator.
This ensures that the 3DMM parameters are optimized for the task of THS.

\section{Joint Training of ReconNet and Generator}
\label{sec:joint_training}

\begin{figure}[t]
	\centering
	\includegraphics[width=11cm]{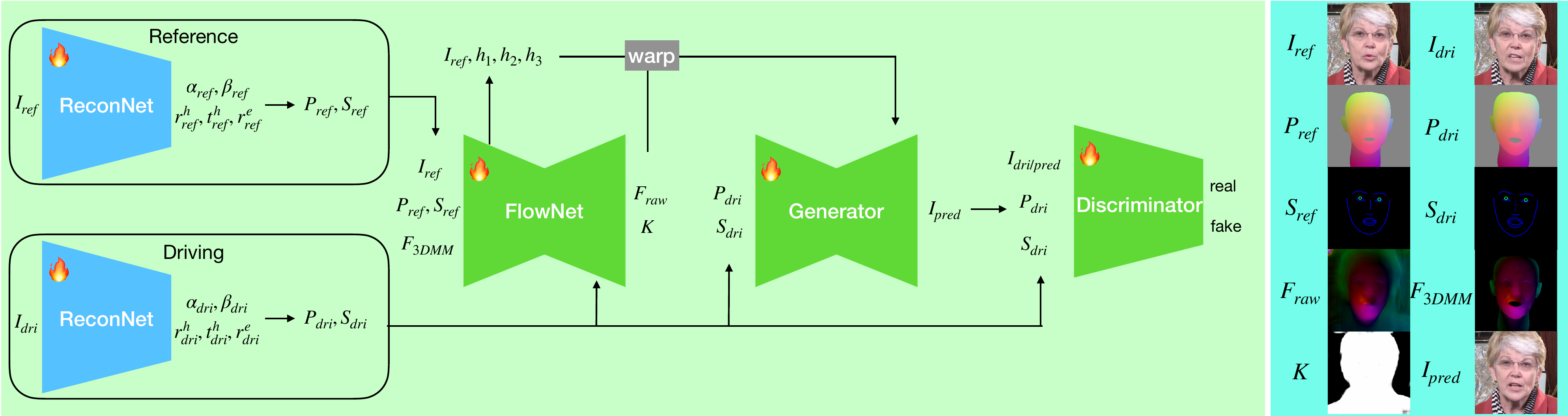}
	\caption{Joint training of ReconNet and Generator. The ReconNet extracts 3DMM parameters from the reference ($I_{ref}$) and the driving ($I_{dri}$) images. These parameters are encoded into $P_{ref/dri}$, $S_{ref/dri}$, and  $F_{3DMM}$. FlowNet takes $I_{ref}$, $P_{ref}$, $S_{ref}$, and $F_{3DMM}$ as inputs to predict the raw flow field $F_{raw}$ and the mask $K$. $I_{ref}$ and the intermediate features of FlowNet ($h_{i}$) are then warped. The Generator takes $P_{dri}$, $S_{dri}$, and the warped features to predict the driving face image $I^{pred}_{dri}$. The Discriminator receives either $I_{pred}$ or $I_{dri}$, along with $S_{dri}$ and $P_{dri}$, to determine whether the image looks realistic.}
    \label{fig:fig1}
\end{figure}

The overview of our joint training scheme is shown in Fig.~\ref{fig:fig1}.
The model architecture is similar to that of HeadGAN~\cite{doukas2021headgan}, but there are important differences that will be discussed below.
For further details, please see the Supplementary Materials (SM).

\subsection{Processing the 3DMM for talking head synthesis}
\label{ssec:encoding_3dmm}

The ICT-FaceKit contains many accessories that are superfluous for our purposes, so we retain only the face and eyeballs.
Because the ICT-FaceKit does not define how the gaze blendshapes (\texttt{eyeLookUp}, \texttt{eyeLookDown}, \texttt{eyeLookLeft}, \texttt{eyeLookRight}) should be coupled to gaze directions, we define these relations ourselves, see SM for pseudocode.

Let $M$ denote the face mesh and $\mathbf{c}$ the projective camera parameters.
To utilize 3DMM for THS, we render the mean face coordinates $P=Renderer(M,\mathbf{c})$, similarly to \cite{zhu2016face,doukas2021headgan}, using PyTorch3D~\cite{ravi2020pytorch3d}.
We also draw sketch $S$ of the face using a two-channel representation: one for the overall face and one for the iris.
We also compute the 2D flow field $F_{3DMM}$ resulting from the changes in 3DMM parameters by calculating how the projected vertex coordinates shift between the reference and target faces, see Fig.~\ref{fig:fig1} for visualization.

\subsection{Pretraining the ReconNet}
\label{ssec:pretraining}
For stable joint training, we pretrain the ReconNet using facial landmarks.
The ReconNet outputs $\alpha$, $\beta$, $r^{h}$, $t^{h}$, $r^{e}$, where $\alpha$ is the identity, $\beta$ is the blendshape, $r^{h}$ and $t^{h}$ are the head rotation and translation, and $r^{e}$ is the eyeball rotation.
The ReconNet is trained to fit the projected 3DMM vertices to 2D landmark labels obtained from off-the-shelf landmark detectors \cite{grishchenko2020attention}.
We use L2 loss for landmarks and apply L2 regularization for identity and blendshapes.
We also include L2 loss between $\alpha$'s extracted from two different images of a person for identity consistency, and constrain the blendshapes to lie between 0 and 1.

\subsection{Joint training of ReconNet and the Generator}
\label{ssec:joint_training}
We modify the feature warping mechanism of HeadGAN, which is unstable.
In the original formulation, FlowNet predicts the mask $K$, which determines where to apply the warping, and the raw flow field $F_{raw}$.
Then, the flow field and the warped image are \href{https://github.com/michaildoukas/headGAN/blob/c1ed8257b447df92d82236d361c72cf929135f1d/models/generator.py#L258}{computed as}:
\begin{align}
    F = K F_{raw}, \quad  I_{warped} = F \star I. \label{eq:unstable_warp}
\end{align}
The multiplication in the first equation is element-wise, and $\star$ denotes the warping operation.
We find that $K$ can collapse to $0$ during training, that is, $F=0$.
We hypothesize that this is because (1) Initially, it is difficult to learn $F$ because the Generator does not produce a meaningful face. (2) As a result, it is easy for the Generator to rely on $S_{dri}$ and $P_{dri}$ for the face shape, and the unwarped features for the face and background textures. (3) Once $K=0$, training gets stuck because the magnitude of $F$ is determined by both $F_{raw}$ (linear activation) and $K$ (sigmoid activation).
We therefore decouple magnitude of flow from $K$ as follows:
\begin{align}
    I_{warped} = K(F_{raw} \star I) + (1-K) I. \label{eq:stable_warp}
\end{align}

For joint training, we add to the losses in Sec.~\ref{ssec:pretraining} the standard photometric losses and the GAN losses.
For spatial locality of mouth blendshapes, we generate face after modifying just the mouth blendshapes, and impose L1 pixel loss on changes that occur outside the mouth region.
To prevent head pose or expression from leaking into the $\alpha$, we randomly swap the identity parameters between the driving and the reference.
To prevent identity from leaking into $\beta$, we randomly shuffle blendshapes within the batch and impose cosine similarity between the ArcFace~\cite{deng2019arcface} embeddings of the original and the generated face.

\section{Application to Lip-Sync}
\subsection{Audio-to-Blendshape}
\label{ssec:audio_to_blendshape}
\begin{figure}[t]
	\centering
	\includegraphics[width=11cm]{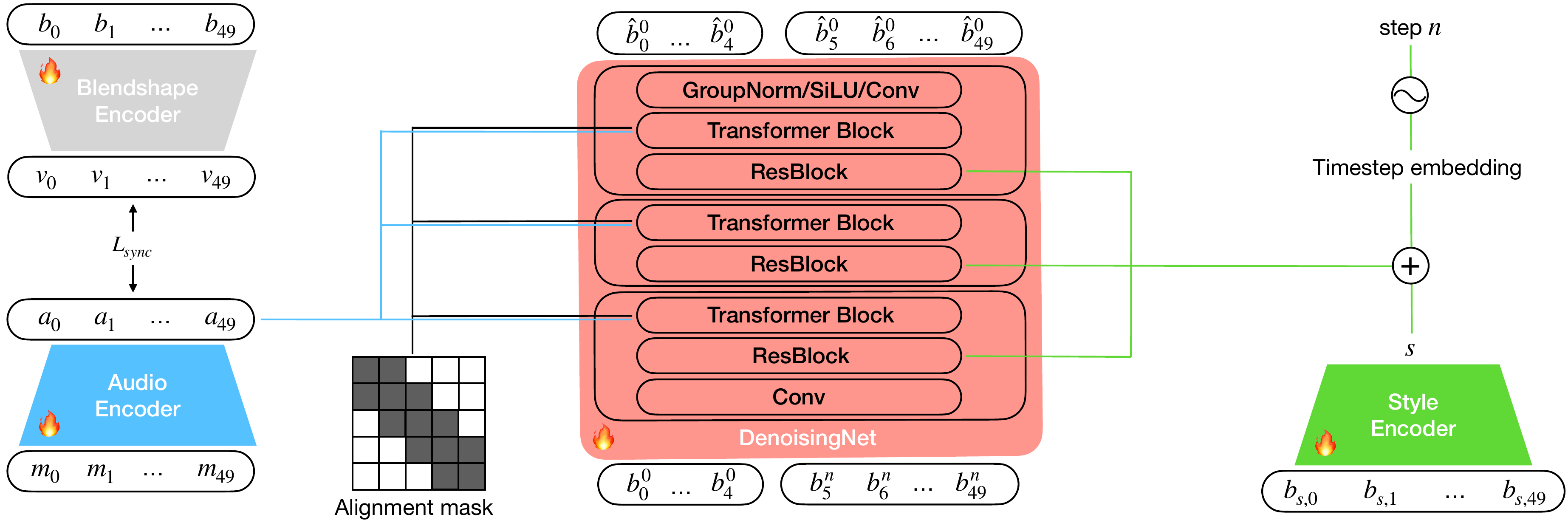}
	\caption{The Audio-to-Blendshape model. The Blendshape Encoder is an auxiliary model for computing the sync-loss, and is not used during inference.}
	\label{fig:fig2}
        \centering
	\includegraphics[width=11cm]{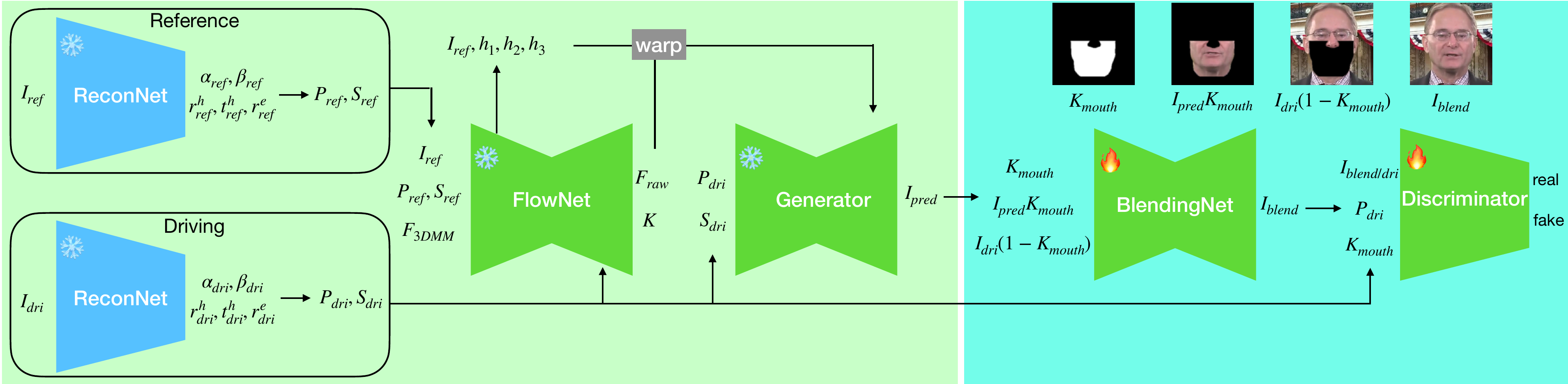}
	\caption{BlendingNet takes the mouth region of the predicted face, the region outside the mouth of the original face, and the mouth mask to predict the blended face.}
	\label{fig:fig3}
\end{figure}
For lip-sync, we train a diffusion model~\cite{park2023said,sun2024diffposetalk} to predict the $35$ mouth blendshapes (shown in Fig.~\ref{fig:fig5}) from audio and talking style embeddings, see Fig.~\ref{fig:fig2}.

First, note that the video is set to $25$ fps and audio is sampled at $16$ kHz.
To compute the audio embeddings, we take $2$s of audio and compute the Mel spectrogram with $80$ frequency bins and a hop size of $200$.
The Mel spectrogram is split into $50$ equally-spaced overlapping chunks $m_i \in \mathbb{R}^{16 \times 80}$.
The Audio Encoder encodes $m_i$ into $a_{i} \in \mathbb{R}^{512}$, $i=0,...,49$.
To capture the speaking style, we sample $50$ consecutive mouth blendshapes $b_{s,i}\in \mathbb{R}^{35}$, $i=0,...,49$ (spanning $2$s) from a different part of the video.
The Style Encoder encodes $b_{s,i}$ into $s\in \mathbb{R}^{512}$.

The DenoisingNet $\mathcal{N}$ transforms noise into mouth blendshapes, conditioned on $a_{i}$ and $s$.
Let $b_{i}$, $i=0,...,49$ be the mouth blendshapes corresponding to $a_{i}$.
The first five blendshapes are used to provide the temporal context to $\mathcal{N}$, enabling it to generate a continuous blendshape sequence.
For the remaining blendshapes, $\mathcal{N}$ is used to transform the noisy blendshapes $b^{n}$ into denoised blendshapes as $\hat{b}^{0} = \mathcal{N}(b_{0:4}, b^{n}_{5:49}, a_{0:49}, n)$.
Here, $n$ is the time step in the denoising process with the convention that $n=0$ corresponds to the original distribution and $n=T$ corresponds to pure noise.

We train using the denoising loss, supplemented by velocity and smoothness losses.
To help train the Audio Encoder, we use the Blendshape Encoder to encode the blendshapes $b_i$ corresponding to $a_i$ into $v_{i}$, and compute the sync-loss between them (see SM for details).

\subsection{Face Blending}
\label{ssec:face_blending}
As will be detailed in Sec.~\ref{ssec:lipsync_pipeline}, a difficulty with lip-syncing is that only the mouth region should be changed.
We approach this problem by first training a neural net that blends the face that is generated from a fixed reference frame into the original video.
For simplicity, we train a UNet $\mathcal{B}$ using the GAN framework (see Fig.~\ref{fig:fig3}).

Let $I_{dri}$ be the original image, $K_{mouth}$ the mouth segmentation map, and $I_{pred}$ the predicted image.
The blended image is given by $I_{blend} = \mathcal{B}(I_{dri}(1-K_{mouth}), K_{mouth}, I_{pred}K_{mouth})$.
The network is trained using photometric and GAN losses.

\begin{figure}[t]
	\centering
	\includegraphics[width=11cm]{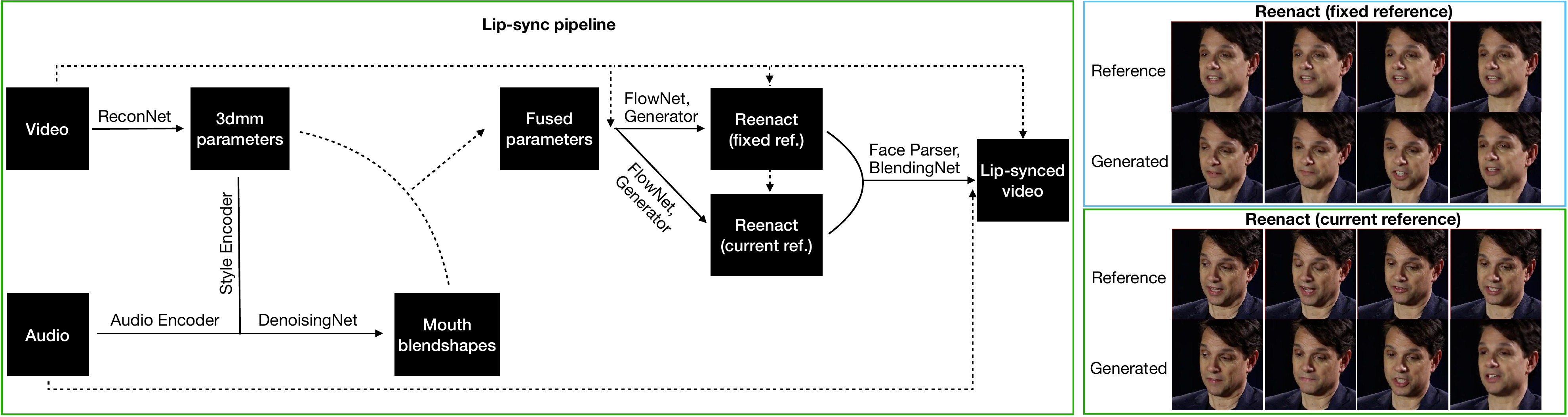}
	\caption{Left: lip-sync pipeline. Right: reenactment with fixed and current references.}
	\label{fig:fig4}
\end{figure}
\subsection{Full Lip-Sync Pipeline}
\label{ssec:lipsync_pipeline}
\begin{figure}
        \centering
	\includegraphics[width=11cm]{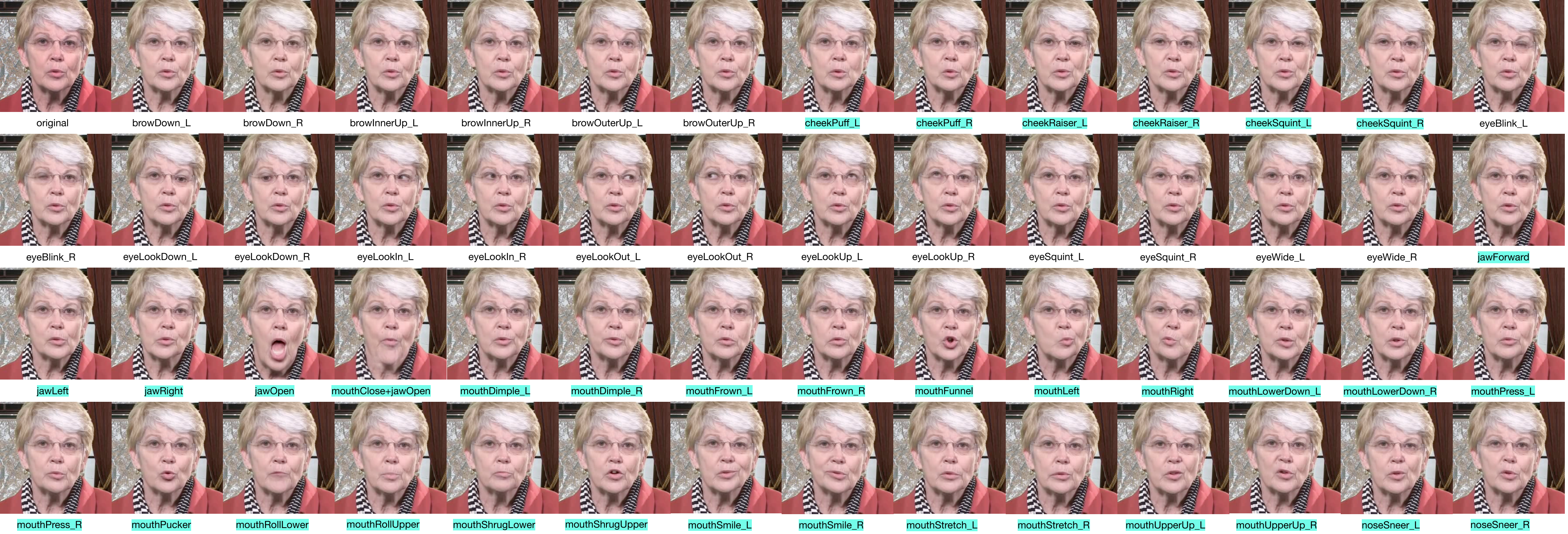}
	\caption{Responses to changes in blendshapes. The mouth blendshapes are highlighted.}
	\label{fig:fig5}
	\centering
	\includegraphics[width=10.5cm]{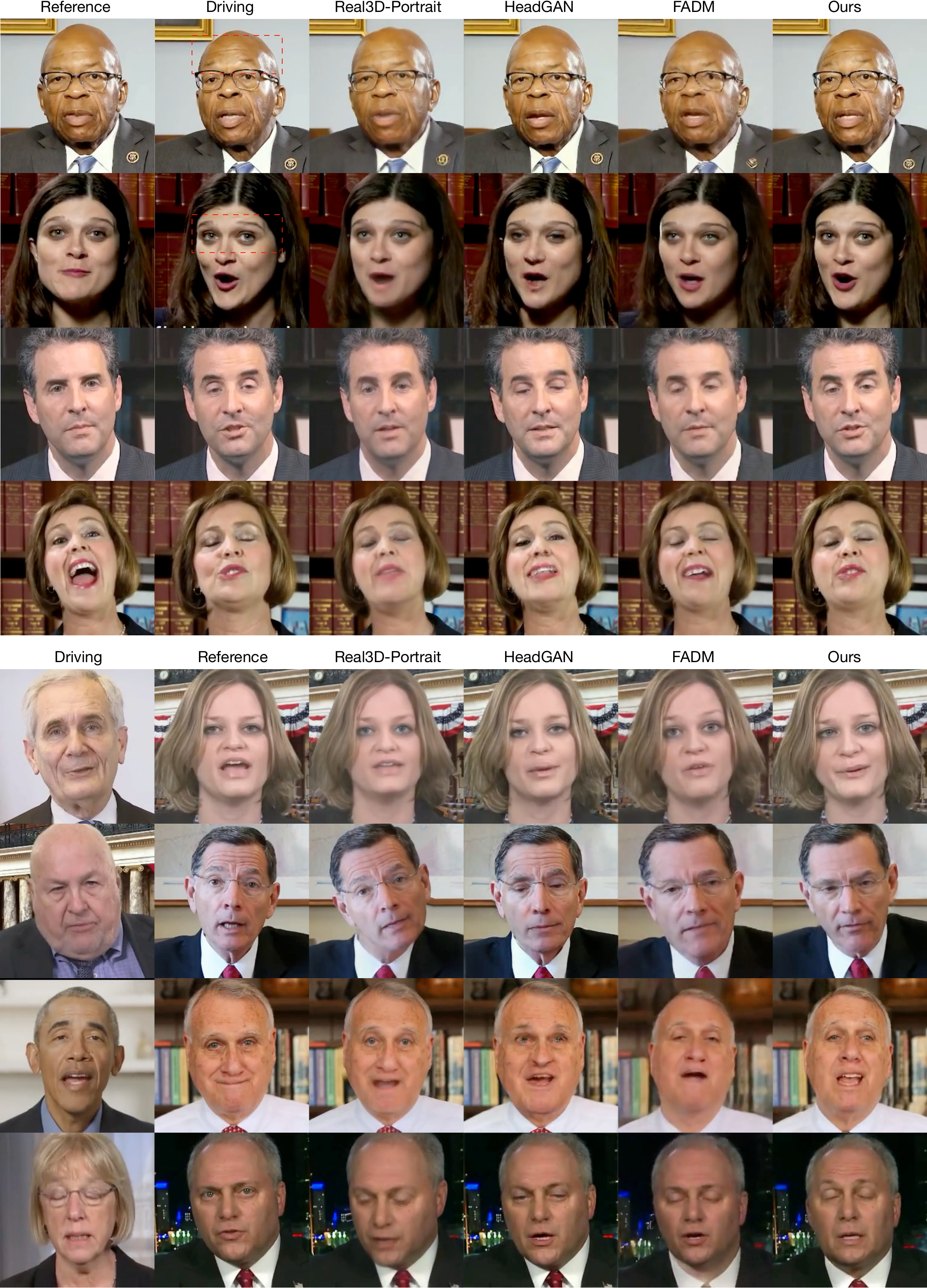}
	\caption{Results of self-reenactment (above) and cross-reenactment (below).}
	\label{fig:fig6}
\end{figure}
Our lip-sync (dubbing) pipeline is shown in Fig.~\ref{fig:fig4}.
We assume that the face to be lip-synced has been detected and that the dubbing audio is provided.
We begin by cropping the face from the video and extracting the 3DMM parameters using the ReconNet.
Next, we extract the style and audio embeddings, which are used to predict denoised mouth blendshapes $\hat{b}$.
These mouth blendshapes $\hat{b}$ are inserted into the 3DMM parameters extracted from the original video.
Using the fused 3DMM parameters, we reenact the (cropped) face twice: (1) with a fixed reference frame to obtain $I_{FF}$, and (2) with the current frame $I_{CF}$ (see Fig.~\ref{fig:fig4}).
We extract the mouth region from $I_{CF}$, and blend $I_{FF}$ into $I_{CF}$.
The blended face is then composited back into the original video.

We now explain the rationale behind generating the face twice to obtain $I_{FF}$ and $I_{CF}$.
As can be expected, simply inpainting the mouth region of the original face using $I_{FF}$ yields unsatisfactory results since the chin contour cannot be modified.
On the other hand, inpainting a large, mouth-shape-agnostic region using $I_{FF}$ can result in noticeable flickering, especially when the neighboring regions of the mask differ between the reference and the current frame.
Using the current frame as the reference also results in unnatural video because the reference frame keeps changing.
We resolve this difficulty by first adjusting the face contour appropriately to obtain $I_{CF}$, and then blending in $I_{FF}$.
This can also enhance video consistency, especially when teeth are visible in the fixed reference frame.

\section{Experiments}
\subsection{Dataset}
We use various publicly available datasets: AVSpeech~\cite{ephrat2018looking}, VoxCeleb2~\cite{chung2018voxceleb2}, CelebV-HQ~\cite{zhu2022celebv}, TalkingHead-1KH~\cite{wang2021one}, and VFHQ~\cite{xie2022vfhq}.
We use the topiq\_nr-face metric in pyiqa \cite{chen2024topiq,pyiqa} to filter out faces with quality lower than $0.4$.
Since audio-visual correlation is important for training the Audio-to-Blendshape model, we restore the sync, and remove data with probability-at-offset $< 0.8$ or offscreen ratio $> 0.354$~\cite{park2024interpretable}.

\subsection{FACS based Blendshapes}
In Fig.~\ref{fig:fig5}, we show that, in many cases, the generated face responds to changes in blendshapes as expected.
However, the model does not respond adequately to blendshapes such as \texttt{cheekPuff}.
We hypothesize that this is due to the use of 2D warping fields and the absence of multi-view consistency during training.

\subsection{Self-reenactment and cross-reenactment}
In Table~\ref{tab:reenactment} and Fig.~\ref{fig:fig6}, we compare our method on the task of facial expression transfer with representative THS methods that utilize 3DMM.
For evaluation, we randomly sample 52 videos from the HDTF dataset, and parts of the VoxCeleb2 dataset (about 1 hour) that are not used for training.
We compute L1, PSNR, SSIM, and FID~\cite{heusel2017gans} to evaluate image quality.
We compute lip-sync metrics LSE-D and LSE-C \cite{prajwal2020lip} to evaluate mouth shape quality, and compute CSIM~\cite{deng2019arcface} to evaluate identity preservation.
Our method outperforms others in most cases, both qualitatively and quantitatively, see SM for more details.
% These datasets have good audio-visual correlations, so we additionally compute the lip-sync quality metrics LSE-D and LSE-C \cite{prajwal2020lip}.
% The results are shown in Table~\ref{tab:reenactment} and Fig.~\ref{fig:fig6}.
% As can be seen, our method outperforms other methods in most cases (see SM for more details).

\subsection{Lip-Syncing}
\begin{figure}[t]
	\centering
	\includegraphics[width=8cm]{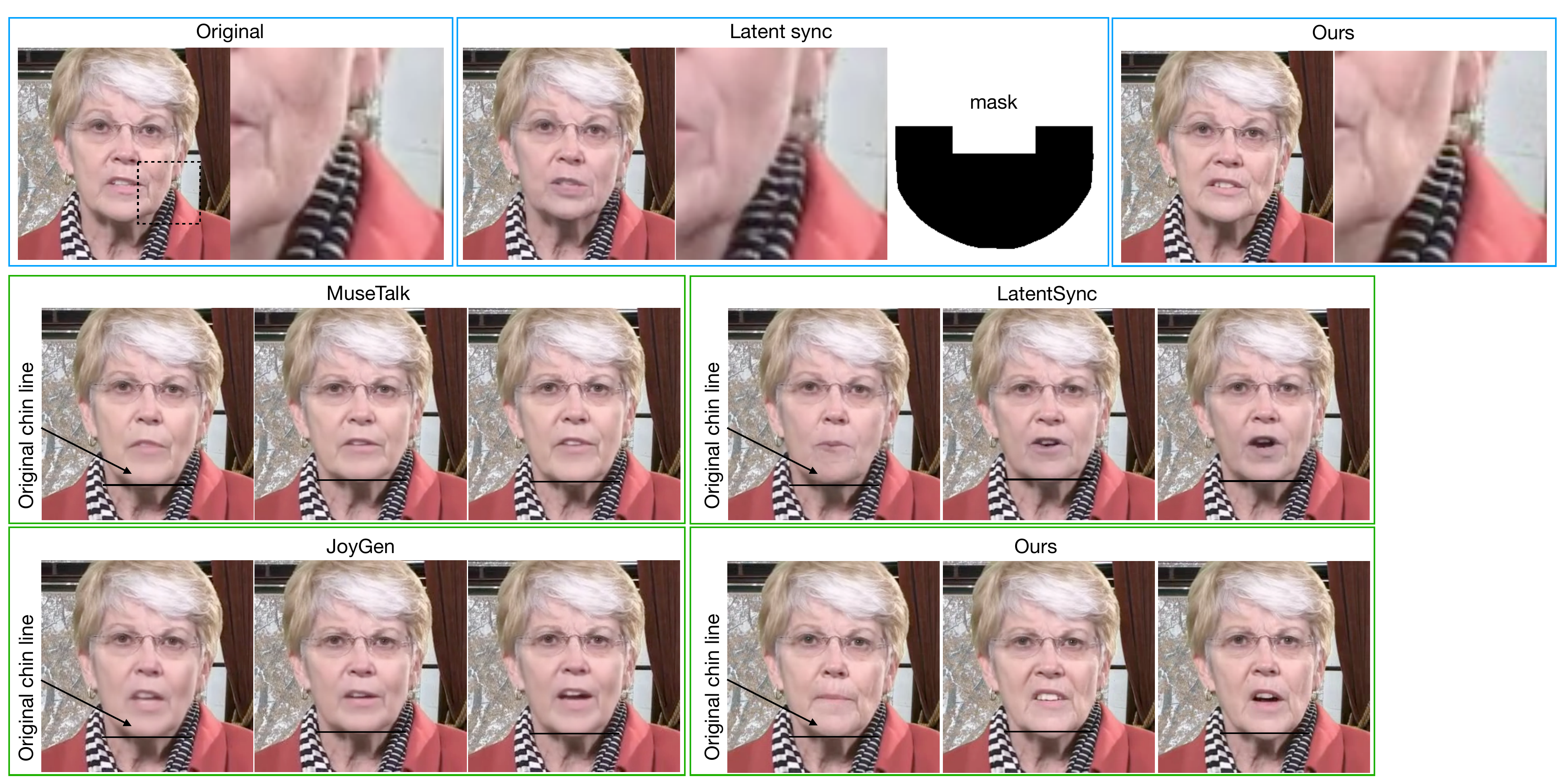}
	\caption{First row: Large mouth mask in LatentSync can cause flickering near the mouth. Second and third rows: MuseTalk and JoyGen cannot change the chin contour, causing face shape to change. Our approach does show these problems.
    }
	\label{fig:fig7}
\end{figure}

In Table~\ref{tab:lipsync}, we compare our method with LatentSync~\cite{li2024latentsync}, MuseTalk~\cite{zhang2024musetalk}, and JoyGen~\cite{wang2025joygen} on the task of lip-sync.
LatentSync outperforms other methods in terms of the LSE-D and LSE-C metrics.
One reason behind this is that LatentSync directly optimizes the sync-loss.
In fact, it achieves better lip-sync metrics even than the original videos, for which LSE-D is $6.8948$ (HDTF) and $7.68$ (VoxCeleb2), and LSE-C is $8.22$ (HDTF) and $6.12$ (VoxCeleb2).

Although our method does not achieve the best LSE-D and LSE-C, it reduces visual artifacts (see Fig.~\ref{fig:fig7}).
LatentSync uses a large, mouth-shape-agnostic mask, and inpaints the masked region.
As can be seen, this can lead to flickering when the visual pattern near the mouth is complex.
In contrast, our approach does not suffer from such artifacts.

Another common artifact of lip-synced videos is the preservation of the original chin contour.
The source of the issue lies in the masking strategy.
For example, MuseTalk and JoyGen inpaint the lower part of the face using a rectangular mask.
This causes the model to associate the chin contour with the mask size.
As a result, the chin line of the original video is preserved in the lip-synced video, which causes lengthening or shortening of the face.
This is not an issue when our lip-sync pipeline is adopted because we modify the original chin contour before blending in the lip-synced face.
To quantitatively evaluate this, we define the metric $\Delta CL = \frac{|y_{2,\textrm{orig}} - y_{2,\textrm{lipsync}}|}{y_{\textrm{2,orig}}-y_{\textrm{1,orig}}}$.
Here, $y_{1}$ and $y_{2}$ are the top and bottom coordinates of face bounding box, and the subscripts orig and lipsync indicate original and lip-synced results, respectively.
As shown in Table~\ref{tab:lipsync} and Fig.~\ref{fig:fig7}, MuseTalk and JoyGen fail to modify the chin line, and the correlation persists in LatentSync as well because the mouth mask cannot completely eliminate information on the original mouth shape.

\begin{table}[t]
    \caption{Evaluation metrics for self-reenactment and \red{cross-reenactment}.
    }
    \centering
    \small
    \setlength{\tabcolsep}{4pt}  % default is usually 6pt
    \begin{tabular}{cccccccc}
        \toprule
         Method        & L1$\downarrow$   & PSNR$\uparrow$ & SSIM$\uparrow$ & FID$\downarrow$     & LSE-D$\downarrow$     & LSE-C$\uparrow$   & CSIM$\uparrow$\\
        \cmidrule(lr){1-8}
        \multicolumn{8}{c}{HDTF dataset} \\
        \cmidrule(lr){1-8}
        Real3D-Portrait \cite{ye2024real3d}& 8.81 & 22.8 & 0.822 & 24.1, \red{48.3} & 8.12, \red{8.42} & 6.71, \red{6.45} & 0.912, \textbf{\red{0.887}} \\
        HeadGAN \cite{doukas2021headgan} & 8.90 & 22.6 & 0.787 & 20.5, \red{47.0} & 7.79, \red{9.10} & 7.06, \red{5.61} & 0.907, \red{0.791} \\
        FADM \cite{zeng2023face} & 7.20 & 25.5 & 0.855 & 16.4, \red{46.4} & 8.12, \red{8.96} & 6.69, \red{5.94} & 0.931, \red{0.845} \\
        Ours            & \textbf{5.18} & \textbf{27.7} & \textbf{0.891} & \textbf{11.3}, \textbf{\red{43.9}} & \textbf{7.02}, \textbf{\red{7.83}} & \textbf{8.12}, \textbf{\red{7.21}} & \textbf{0.949}, \red{0.833} \\
        \cmidrule(lr){1-8}
        \multicolumn{8}{c}{VoxCeleb2 dataset} \\
        \cmidrule(lr){1-8}
        Real3D-Portrait \cite{ye2024real3d} & 12.2 & 21.1 & 0.799 & 17.1, \textbf{\red{19.5}} & 9.00, \red{9.42}  & 4.66, \red{4.22} & 0.832, \textbf{\red{0.808}} \\
        HeadGAN \cite{doukas2021headgan} & 12.1 & 21.4 & 0.742 & 13.1, \red{24.9} & 8.51, \red{9.26}  & 5.11, \red{4.22} & 0.795, \red{0.442} \\
        FADM \cite{zeng2023face} & 9.97 & 23.9 & 0.817 & 9.97, \red{23.1} & 9.07, \red{9.88}  & 4.46, \red{3.63} & 0.882, \red{0.624} \\
        Ours            & \textbf{7.42} & \textbf{25.7} & \textbf{0.852} & \textbf{7.19}, \red{24.9} & \textbf{7.90}, \textbf{\red{8.62}}  & \textbf{5.94}, \textbf{\red{5.12}} & \textbf{0.912}, \red{0.652} \\
        \bottomrule
    \end{tabular}
    \label{tab:reenactment}
    \caption{Evaluation metrics for lip-sync with original and \red{different audio}.
    }
    \centering
    \small
    \begin{tabular}{cccccc}
        \toprule
         Method        & FID$\downarrow$     & LSE-D$\downarrow$  & LSE-C $\uparrow$ & CSIM$\uparrow$ & $\Delta$CL$\uparrow$ \\
        \cmidrule(lr){1-6}
        \multicolumn{6}{c}{HDTF dataset} \\
        \cmidrule(lr){1-6}
        LatentSync \cite{li2024latentsync} & 4.05, \red{11.4} &  \textbf{5.78}, \textbf{\red{6.00}}  & \textbf{9.76}, \textbf{\red{9.41}} & 0.967, \red{0.962} & 1.04, \red{1.26} \\
        MuseTalk \cite{zhang2024musetalk} & 4.76, \red{11.9} &  8.30, \red{10.8}  & 6.64, \red{4.09} & 0.967, \red{0.965} & 0.882, \red{0.957} \\
        JoyGen \cite{wang2025joygen} & 8.87, \red{18.0} &  7.19, \red{8.90}  & 7.89, \red{5.90} & 0.971, \textbf{\red{0.968}} & 0.832, \red{0.926} \\
        Ours blended    & \textbf{3.51}, \textbf{\red{10.5}} &  8.26, \red{8.99}  & 6.58, \red{5.80} & \textbf{0.972}, \red{0.967} & \textbf{1.08}, \textbf{\red{1.32}} \\
        \cmidrule(lr){1-6}
        % \multicolumn{6}{c}{HDTF dataset, audio based reenactment} \\
        % \cmidrule(lr){1-6}
        % Ours            & 11.6, \red{17.6} &  8.25, \red{8.90}  & 6.61, \red{5.92} & 0.948, \red{0.947} & 1.14, \red{1.35} \\
        % Real3D-Portrait & 24.7, \red{28.0} &  7.21, \red{7.51}  & 7.66, \red{7.36} & 0.912, \red{0.911} & 1.82, \red{1.98} \\
        % \cmidrule(lr){1-6}
        \multicolumn{6}{c}{VoxCeleb2 dataset} \\
        \cmidrule(lr){1-6}
        LatentSync \cite{li2024latentsync} & 2.15, \red{3.22} & \textbf{6.41}, \textbf{\red{6.84}} & \textbf{7.94}, \textbf{\red{7.21}} & 0.964, \red{0.953} & 1.04, \red{1.19} \\
        MuseTalk \cite{zhang2024musetalk} & 2.93, \red{3.81} & 8.58, \red{10.7} & 5.11, \red{2.88} & 0.918, \red{0.910} & 1.03, \red{1.05} \\
        JoyGen \cite{wang2025joygen} & 5.14, \red{6.74} & 7.72, \red{9.54} & 6.10, \red{4.16} & 0.938, \red{0.927} & 1.02, \red{1.06} \\
        Ours blended    & \textbf{1.91}, \textbf{\red{2.85}} & 9.57, \red{9.57} & 3.93, \red{3.93} & \textbf{0.965}, \textbf{\red{0.959}} & \textbf{1.17}, \textbf{\red{1.33}} \\
        % \multicolumn{6}{c}{VoxCeleb2 dataset, audio based reenactment} \\
        % \cmidrule(lr){1-6}
        % Ours            & 7.55, \red{8.55} & 9.01, \red{9.47} & 4.63, \red{4.07} & 0.912, \red{0.910} & 1.32, \red{1.45} \\
        % Real3D-Portrait & 16.7, \red{17.3} & 8.17, \red{8.44} & 5.62, \red{5.21} & 0.836, \red{0.836} & 2.16, \red{2.23} \\
        \bottomrule
    \end{tabular}
    \label{tab:lipsync}
\end{table}
\section{Conclusion}
Contrary to common belief, we have shown that 3DMM remains a suitable representation for the purpose of THS.
The limitations of previous approaches that rely on 3DMM stem from poor extraction of 3DMM parameters, rather than from limitations of 3DMM itself.
Using the FACS-based blendshapes of ICT-FaceKit, we have also proposed a lip-sync pipeline that allows modification of the chin contour and reduces flickering near the mouth.

The fact that warping-field-based approaches to THS are highly compatible with the joint learning of ReconNet has important implications.
For example, our method can readily be generalized to approaches using 3D warping fields such as face-vid2vid~\cite{wang2021one} by binding 3D keypoints to vertices of the 3DMM.
It would also be interesting to explore whether keypoints can be bound to the 3DMM to control the tongue movement, or whether the 3DMM can be supplemented with additional latent vectors that capture fine-grained facial details, such as wrinkles.
We leave the answers to these questions for future research.

\bibliography{egbib}

\clearpage
\appendix
% \section{Notation}
% \begin{table}[h]
%     \caption{Notation}
%     \centering
%     \begin{tabular}{ll}
%         \toprule
%         $I$ & image \\
%         $m$ & mel spectrogram \\
%         $\alpha$ & identity \\
%         $\beta$ & expression \\
%         $r$ & head or eye rotation \\
%         $t$ & head translation \\
%         $c$ & camera parameters \\
%         $P$ & rendered image of mean face coordinates \\
%         $M$ & mesh \\
%         $Renderer$ & renderer \\
%         $F$ & flow field \\
%         $S$ & sketch of face \\
%         $\mathcal{R}$ & recon-net \\
%         $\mathcal{F}$ & flow net \\
%         $\mathcal{G}$ & generator \\
%         $\mathcal{D}$ & discriminator \\
%         $\mathcal{A}$ & audio encoder \\
%         $\mathcal{S}$ & style encoder \\
%         $\mathcal{N}$ & denoising network \\
%         $\mathcal{B}$ & face blending network \\
%         \bottomrule
%     \end{tabular}
%     \label{tab:notation}
% \end{table}
% \clearpage
\section{3DMM geometry}
\begin{figure}
	\centering
	\includegraphics[width=8cm]{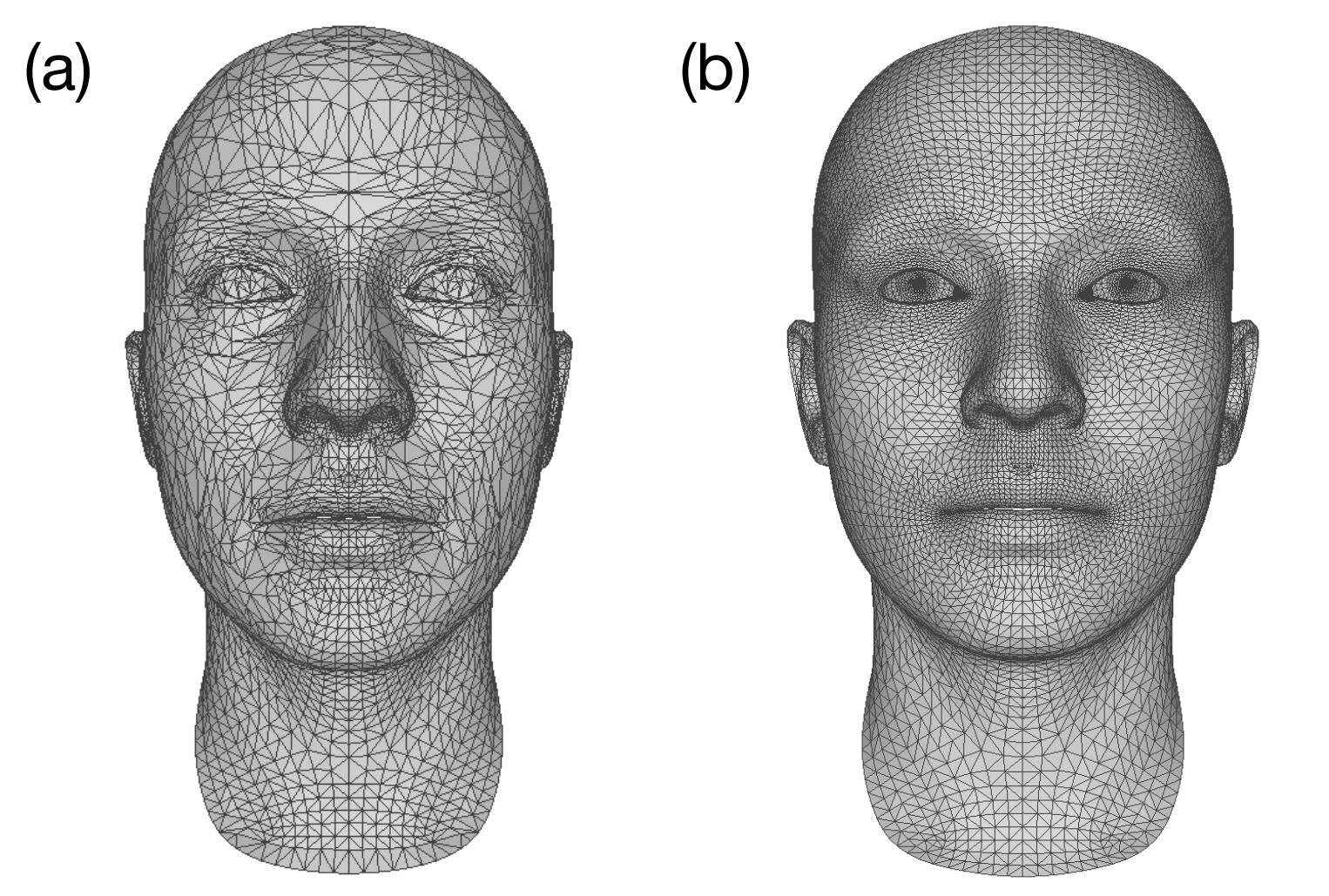}
	\caption{(a) Simplified geometry (b) Original geometry after triangulation.}
    \label{fig:fig8}
\end{figure}
We keep only the face and the eyeballs of the ICT-FaceKit.
After triangulation, the ICT-FaceKit has 12,549 vertices and 24,852 faces.
We further simplify the mesh to have 5,099 vertices and 9,857 faces (as a comparison, the FLAME face mesh has 5,023 vertices and 9,976 faces).
The mesh is simplified by using the quadratic mesh error \cite{garland1997surface}, except that we average the error over 50 randomly generated expressions, and decimate vertices symmetrically.
The result is shown in Fig.~\ref{fig:fig8}.
While this simplification is not necessary, it reduces unnecessary compute.
Note that we use all of the blendshapes and the first 50 identity parameters of the ICT-FaceKit.

We next give pseudocode for encoding the 3DMM (see discussion in Sec.~\ref{ssec:encoding_3dmm} and visualization in Fig.~\ref{fig:fig1}) using PyTorch and PyTorch3D.
$P$ is computed as follows:
\begin{lstlisting}
#  B: batch size, N_V: number of vertices, N_T: number of triangles
#  v_f: vertices in facial coordinates of shape [B, N_V, 3]
#  mesh: face mesh transformed to world coordinates
#  rasterizer: pytorch3d rasterizer
fragments = rasterizer(mesh)
pix_to_face = fragments.pix_to_face # [B, H, W, 1]
bary_coords = fragments.bary_coords.unsqueeze(-1) # [B, H, W, 1, 3, 1]
foreground = (pix_to_face >= 0).permute(0, 3, 1, 2) # [B, 1, H, W]
packed_faces = mesh.faces_packed() # [B*N_T, 3]
packed_verts = mesh.verts_packed() # [B*N_V, 3]
pix_to_verts_idx = packed_faces[pix_to_face] # [B, H, W, 1, 3]
P = v_f.view(-1, 3)[pix_to_verts_idx] * bary_coords
P = P.sum(axis=-2).squeeze(-2).permute(0,3,1,2)
P = P * foreground # [B, 3, H, W]
\end{lstlisting}
$F_{3DMM}$ is computed as follows:
\begin{lstlisting}
#  camera: pytorch3d camera
#  d: driving, r: reference
#  packed_verts, pix_to_verts_idx, packed_proj_verts, 
#  bary_coords, foreground are computed as before
packed_proj_verts_d = camera.transform_points_screen(packed_verts_d)
packed_proj_verts_r = camera.transform_points_screen(packed_verts_r)
pix_to_verts_d = (packed_proj_verts_d.view(-1,3)[pix_to_verts_idx_d] * bary_coords_d.unsqueeze(-1)).sum(axis=-2).squeeze(-2)
pix_to_verts_sd = (packed_proj_verts_s.view(-1,3)[pix_to_verts_idx_d] * bary_coords_d.unsqueeze(-1)).sum(axis=-2).squeeze(-2)
F_3dmm = (pix_to_verts_sd - pix_to_verts_d) * foreground_d
\end{lstlisting}

The pseudocode for computing the gaze blendshapes from gaze direction for the right eyeball is as follows:
\begin{lstlisting}
# rot_matrix: rotation matrix for right eyeball of shape [B, 3, 3]
# th_max, tv_max: maximum horizontal and vertical gaze angles
d = rot_matrix[:, 2]  # gaze direction
th = torch.atan2(d[:, 0], d[:, 2])  # horizontal angle
tv = torch.atan2(d[:, 1], d[:, 2])  # vertical angle
norm_h = (th / th_max)  # normalize horizontal angle
norm_v = (tv / tv_max)  # normalize vertical angle
eye_look_in = norm_h.clamp(0, 1)
eye_look_out = -norm_h.clamp(-1, 0)
eye_look_up = norm_v.clamp(0, 1)
eye_look_down = -norm_v.clamp(-1, 0)
\end{lstlisting}
Note that $x$-axis points to the right, $y$-axis points upwards, and $z$-axis points out of the plane in Fig.~\ref{fig:fig8}.
Therefore, when all gaze related blendshapes are zero, the gaze direction points towards the $z$-axis.
The computation for the left eyeball is similar.
Note that the maximum gaze angles in the horizontal direction (\texttt{th\_max}) and vertical direction (\texttt{tv\_max}) are typically set to be around $30^{\circ}$~\cite{zhang2024refa}.

\section{Further details on the architecture}
\begin{figure}
	\centering
	\includegraphics[width=12cm]{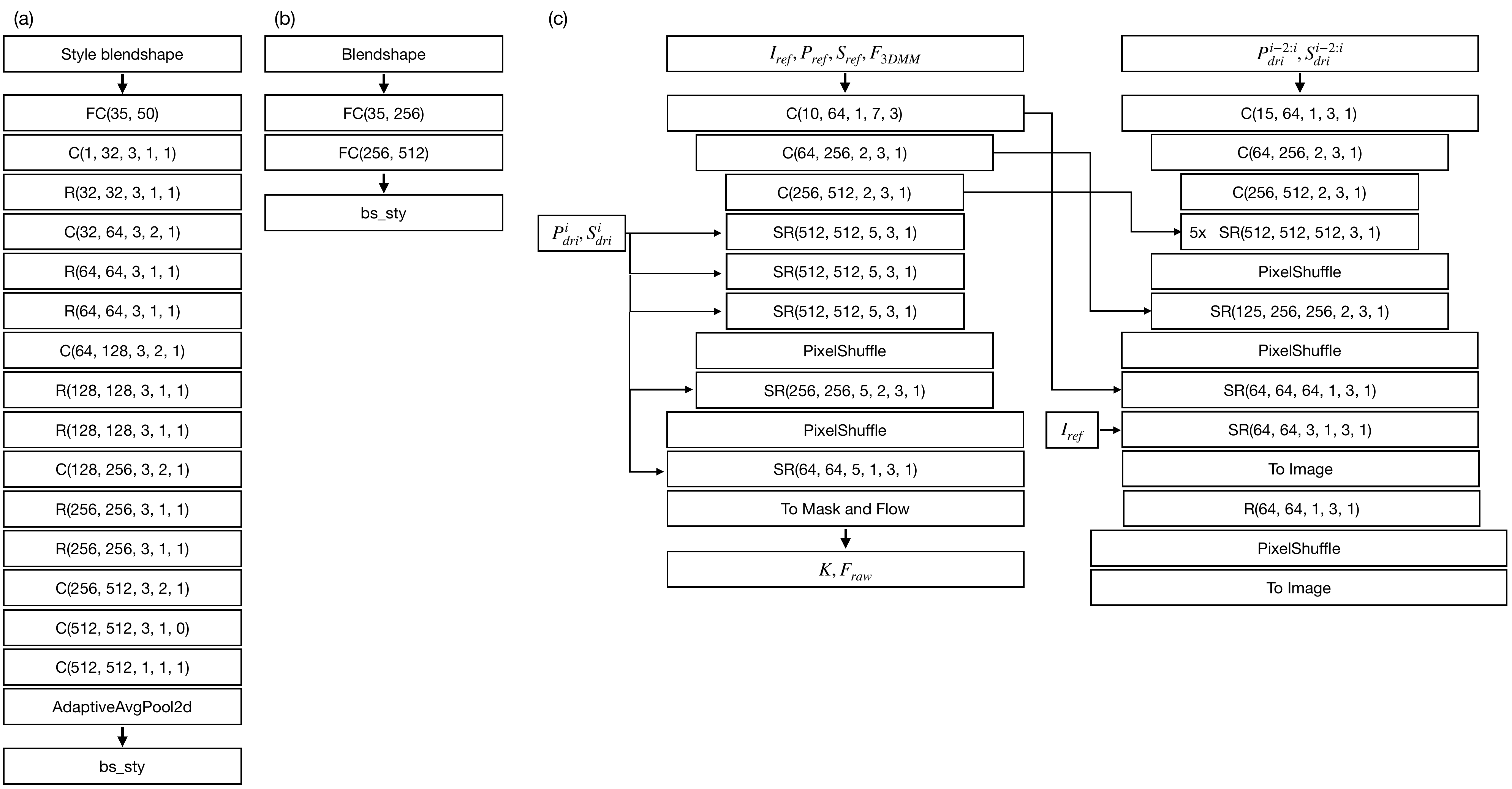}
	\caption{(a) Style Encoder. (b) Blendshape Encoder. (c) FlowNet and Generator. \texttt{FC} represents fully-connected layer with Leaky ReLU activation. \texttt{C} represents simple ConvBlock with group normalization and Leaky ReLU activation. \texttt{R} represents ResBlock. \texttt{SR} represents SPADE ResBlock. The numbers in the parenthesis in \texttt{C} and \texttt{R} are input channels, output channels, kernel size, stride, and padding. 
    The numbers in the parenthesis in \texttt{SR} are input channels, output channels, conditioning channels, kernel size, stride, and padding. 
    In (a), the output is reshaped to [B, 1, 50, 50] after \texttt{FC}.
    In (c), \texttt{To Image} is a simple ConvBlock with sigmoid activation. \texttt{To Mask and Flow} is also a simple ConvBlock, where \texttt{Flow} has linear activation and \texttt{Mask} has sigmoid activation.}
    \label{fig:fig9}
\end{figure}

\textbf{ReconNet}: 
We use ResNet-50 backbone~\cite{he2016deep}, except that the head predicts $\alpha \in \mathbb{R}^{50}$; $\beta \in \mathbb{R}^{55}$; $r^{h}, r^{e} \in \mathbb{R}^{12}$; $t_{head} \in \mathbb{R}^{3}$.
Note that we use a 6D representation of rotation~\cite{zhou2019continuity}.
The input image, $P$, and $S$ are all at $224\times 224$ resolution.

\noindent
\textbf{FlowNet, Generator}.
The architecture of FlowNet is adapted from HeadGAN with the following changes:
\begin{itemize}
    \item InstanceNorm~\cite{ulyanov2016instance} is replaced by GroupNorm~\cite{wu2018group}.
    \item Input channel is 10 ($I$, $P$, $S$, $F_{3DMM}$).
    \item Output channel of first convolution block is $64$, instead of $32$.
\end{itemize}
Note that the maximum number of channels is kept at $512$, and that input and output resolution are $256$.
The architecture of Generator is adapted from HeadGAN with the following changes:
\begin{itemize}
    \item InstanceNorm is replaced by GroupNorm.
    \item Input channel is $15$ (three consecutive $P$ and $S$).
    \item Output channel of first convolution block is $64$, instead of $32$.
    \item The number of SPADE layers at the lowest resolution is increased from $1$ to $5$.
    \item We remove AdaIN blocks that use audio signals.
\end{itemize}
Note that max channel is kept at $512$, and that the input and output resolution is $256$.
See Fig.~\ref{fig:fig9} (c).

\noindent
\textbf{Discriminator}.
The architecture of Discriminator is adopted from HeadGAN, which is similar to that in \cite{park2019semantic}, except for the following changes:
\begin{itemize}
    \item InstanceNorm is changed to GroupNorm
    \item Input channel size is $8$ ($I$, $P$, $S$)
\end{itemize}
We use three discriminators, one for the full image, one for the eyes, and one for the mouth.

\noindent
\textbf{AudioEncoder}.
The audio encoder in the Audio-to-Blendshape module is the same as the audio encoder in Wav2Lip~\cite{prajwal2020lip}.

\noindent
\textbf{Audio-to-blendshape}.
The style encoder in the Audio-to-Blendshape module is shown in Fig.~\ref{fig:fig9} (a).
The blendshape encoder for computing sync-loss is shown in Fig.~\ref{fig:fig9} (b).
The denoising net follows the simple UNet-style convolutional network with cross attention in SAiD~\cite{park2023said}.

\noindent
\textbf{BlendingNet}.
The BlendingNet is a simple UNet~\cite{ronneberger2015u,isola2017image}.
The output of UNet ($I_{pred}$) is blended into the original image as follows:
\begin{equation}
    I_{blend} = I_{orig}  (1-K_{mouth}) + I_{pred} K_{mouth}
\end{equation}

\section{Further details on training}
\subsection{Pretraining the ReconNet}
For faster convergence, we initialize the backbone of ReconNet with a pretrained backbone~\cite{deng2019accurate}.
We use L2 loss for landmark $\mathcal{L}_{lm} = \lVert l_{label} - l_{3DMM}\rVert_{2}^{2}$ and the L2 regularization $\mathcal{L}_{reg} = \lVert \alpha\rVert_{2}^2 + \lVert \beta \rVert_{2}^2$.
Note that $l_{label}$ consists of $68$ standard facial landmarks supplemented by $10$ additional landmarks for the iris, which are extracted using MediaPipe~\cite{grishchenko2020attention}.
We also include identity consistency loss $\mathcal{L}_{id} = \lVert \alpha_1 - \alpha_2 \rVert_{2}^2$, where $1$ and $2$ indicate identity parameters extracted from the same video but from different frames.
Since blendshapes are designed to lie between $0$ and $1$, we penalize deviations using $\mathcal{L}_{constraint} = \textrm{mean}(\textrm{max}(\textrm{abs}(\beta - 0.5), 0.5) - 0.5)$,
where max and abs are element-wise operations.
The full ReconNet loss $\mathcal{L}_{recon}$ is a sum of these losses, with the weights given by
$\lambda_{lm} = 80$, $\lambda_{reg} = 0.025$, $\lambda_{id}=0.025$, $\lambda_{constraint}=0.2$. % $\lambda_{calibration}=0.4$
We train for $100,000$ steps using the Adam~\cite{kingma2014adam} optimizer with batch size $128$, a learning rate of $0.0001$, and betas $(0.9, 0.999)$.
Training takes about 1.5 days on a single NVIDIA A100 GPU.

\subsection{Joint training of ReconNet and Generator}
We use the VGG perceptual loss~\cite{ledig2017photo,simonyan2014very} $\mathcal{L}_{perceptual}$ and L1 pixel loss $\mathcal{L}_{pixel}$ on the foreground regions between the ground truth and predicted images. 
We use the GAN hinge loss~\cite{lim2017geometric} $\mathcal{L}_{GAN}$, along with feature matching loss~\cite{salimans2016improved} $\mathcal{L}_{FM}$.
The mask $K$ is guided using L1 loss $\mathcal{L}_{FG}$ to predict the foreground region \cite{lin2022robust}.
We use VGG perceptual loss $\mathcal{L}_{perceptual}^w$ and masked L1 pixel loss $\mathcal{L}_{pixel}^w$ between the warped image and the ground truth image.
As auxiliary losses, we use L1 loss between $F_{3DMM}$ and $F$ on the face region to encourage the predicted flow to be consistent with 3DMM.
To enforce spatial locality, we detect the mouth region and use L1 loss $\mathcal{L}_{locality}$ to demand that when mouth blendshapes are changed by shuffling the mouth blendshapes within the batch, the generated image does not change outside the mouth region.
To disentangle expression from identity and head pose, we shuffle expressions inside the batch and impose the cosine similarity loss $\mathcal{L}_{arc}=1-\hat{z}_{arc} \cdot \hat{z}_{arc}'$ between the original face and the reenacted face, where $\hat{z}_{arc}$ is the unit-norm ArcFace \cite{deng2019arcface} embedding.
We also randomly shuffle identity parameters between reference and driving images to prevent expression and head pose from leaking into the identity parameters.
The loss weights are as follows: 
$\lambda_{perceptual}=10$, $\lambda_{GAN}=1$, $\lambda_{FM}=10$, $\lambda_{pixel}=50$, $\lambda_{perceptual}^w=10$, $\lambda_{pixel}^{w}=50$, $\lambda_{FG}=10$, $\lambda_{locality}=20$, $\lambda_{arc}=1$.
We also add L2 smoothness loss $\mathcal{L}_{smooth}$ with $\lambda_{smooth}=0.025$, so that blendshapes, head translations, head rotations, eye rotations between consecutive frames do not change too much.

We train with the Adam optimizer with an initial learning rate of $0.0002$, $\beta=(0.5,0.999)$, and batch size of $80$. 
We train for $150,000$ steps, where we reduce the learning rate by half starting from $50,000$ steps with an interval of $15,000$ steps.
Training takes about 4 days on 8 NVIDIA A100 GPUs.

\subsection{Audio-to-Blendshape}
We predict the denoised blendshapes, so the loss is $\mathcal{L}_{simple} = \lVert \hat{b}_{0:49}^{0} - b_{0:49}^{0} \rVert^2_2$.
We also include the velocity loss $\mathcal{L}_{vel} = \lVert (b^{0}_{1:49} - b^{0}_{0:48}) - (\hat{b}^{0}_{1:49}-\hat{b}_{0:48})\rVert^2_2$, and the smoothness loss $\mathcal{L}_{smooth} = \lVert \hat{b}^{0}_{2:49} - 2\hat{b}^{0}_{1:48} + \hat{b}^{0}_{0:47}\rVert^2_2.$
To help train the Audio Encoder, we also include the sync-loss as follows.
Recall from Sec.~\ref{ssec:audio_to_blendshape} that the Mel spectrograms are encoded into $a_{i} \in \mathbb{R}^{512}$, $i=0,...,49$.
We encode $b_{i}$, $i=0,...,49$ corresponding to $a_{i}$ into $v_{i} \in \mathbb{R}^{512}$ using a shallow network (Blendshape Encoder in Fig.~\ref{fig:fig2}).
The sync-loss is given by
\begin{equation}
    \mathcal{L}_{sync} = -\frac{1}{50\times 2}\sum_{i=0}^{49}\log \frac{e^{\phi(v_{i}, a_{i})}}{\sum_{j \notin \{i-1,i+1\} } e^{\phi(v_i, a_{j})}} - (a \leftrightarrow v)
\end{equation}
During training, we randomly drop the audio signal with probability $0.1$, the previous audio with probability $0.5$, and the previous blendshape sequence with probability $0.5$.
The loss weights are $\lambda_{simple}=1$, $\lambda_{vel}=7.5$, $\lambda_{smooth}=1$, $\lambda_{sync}=1$.
The total number of denoising steps $T$ is set to $1000$.
During inference, we use $50$ denoising steps.
We train using the AdamW~\cite{loshchilov2017decoupled} optimizer with betas $(0.9, 0.999)$, a learning rate of $0.0001$, and a batch size of $128$ for $500,000$ steps.
Training takes about 2 days on a single NVIDIA A100 GPU.
Note that during inference, we use classifier free guidance~\cite{ho2022classifier} with the guidance scale fixed to 1.2 for audio and style.

\subsection{BlendingNet}
We train using $\mathcal{L}_{perceptual}$, $\mathcal{L}_{pixel}$, $\mathcal{L}_{GAN}$, $\mathcal{L}_{FM}$ with weights $\lambda_{perceptual}=10$, $\lambda_{pixel}=50$, $\lambda_{GAN}=1$, and $\lambda_{FM}=10$.
We use the Adam optimizer with $\beta=(0.5, 0.999)$, a learning rate of $0.0002$, and a batch size of $80$.
Training is performed for $150,000$ steps, where the learning rate is halved starting from $50,000$ steps with an interval of $15,000$ steps.
Training takes about 7 days on 4 NVIDIA A100 GPUs.

\subsection{Inference Time}
When using a single NVIDIA 3090 GPU, the full lip-sync pipeline typically takes around 9 times the duration of the video, where batch size is set to $8$ except for the Audio-to-Blendshape model, for which we set the batch size to $1$.
This includes loading times, the caching time (we cache results such as $P$ and $S$, $I_{FF}$, $I_{CR}$ to the hard drive), and the time required to save the final result.
Note that mouth mask is created using the face parser provided in FaRL~\cite{zheng2022general}.

\section{Ablation studies}
\begin{figure}
	\centering
	\includegraphics[width=10cm]{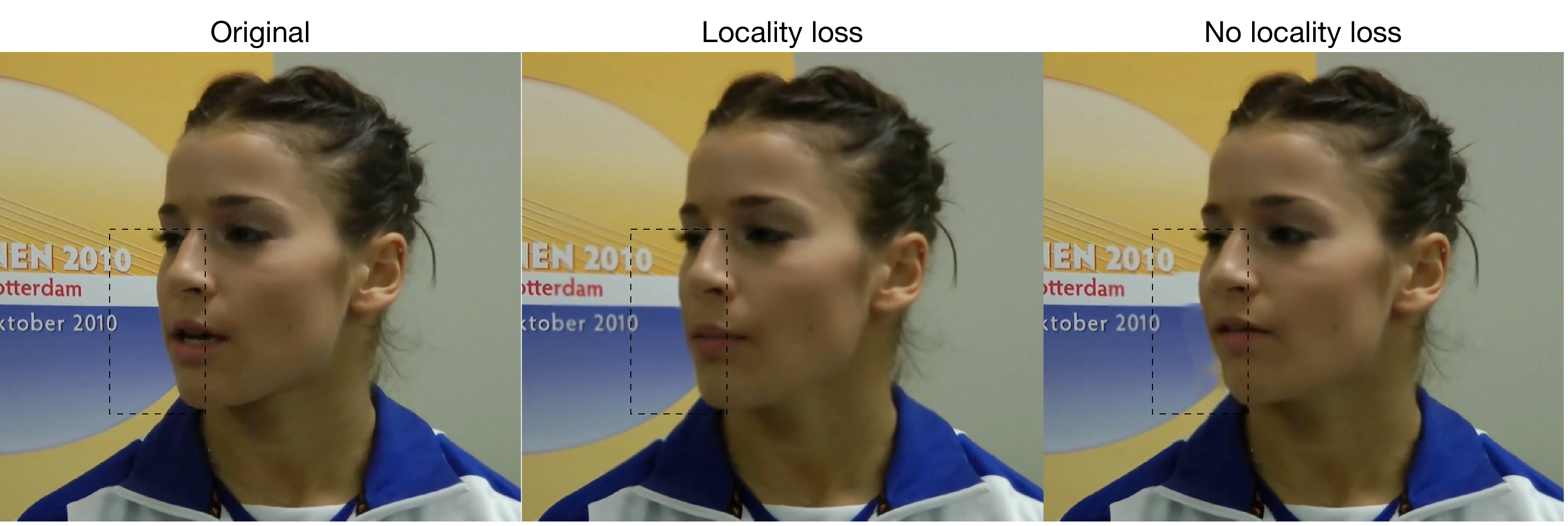}
	\caption{The effect of locality loss.
    When locality loss is not imposed, visible flickering appears in the background region, which is indicated with a dashed box.
    This is significantly reduced when the locality loss is imposed.}
    \label{fig:fig10}
\end{figure}

\textbf{Locality loss.}
The locality loss is necessary for natural lip-sync. 
Recall that the locality loss demands that the areas outside the mouth does not change when the mouth blendshapes are changed:
\begin{equation}
\mathcal{L} = \lVert (1-K_{mouth})(I_{pred}^{m} - I_{ref}) \rVert_{1}.
\end{equation}
Here, $I_{pred}^{m}$ is the predicted image when mouth blendshapes are changed, and $I_{ref}$ is the reference image.
Without the locality loss, there can be flickering in the surrounding region when the current frame is used as the reference to change the chin contour.
This is shown in Fig.~\ref{fig:fig10}.

\begin{figure}
	\centering
	\includegraphics[width=10cm]{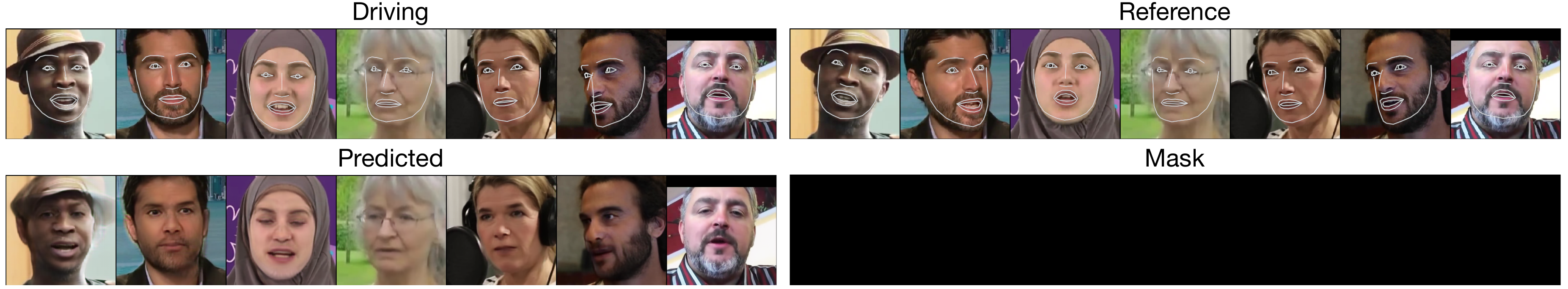}
	\caption{Importance of properly parameterizing the warping operation can be seen from this example run where the mask $K$ collapses to zero.
    The face sketch $S$ drawn using the 3DMM parameters predicted by the ReconNet is overlaid for reference.
    }
    \label{fig:fig11}
\end{figure}

\noindent
\textbf{Parametrization of warping.}
The parametrization of the warping operation in Eq.~\ref{eq:stable_warp} is crucial for convergence.
We have also tried detaching the gradient (stopgrad) for the mask $K$ in Eq.~\ref{eq:unstable_warp}:
\begin{equation}
    F = \textrm{sg}(K)F_{raw}, \quad I_{warped}=F \star I. \label{eq:detach_warp}
\end{equation}
We find that this parametrization is more stable than Eq.~\ref{eq:unstable_warp}, but even Eq.~\ref{eq:detach_warp} can cause $K$ to collapse to zero.
An example run where we observed this behavior is shown in Fig.~\ref{fig:fig11}.
Because $K=F=0$, features are not warped, and the Generator relies on $S_{dri}, P_{dri}$ for the face shape, and on unwarped features for face texture.
This is neither intended nor helpful for THS.
We have not observed this instability when Eq.~\ref{eq:stable_warp} is used.

\noindent
\textbf{Pretraining the ReconNet.}
We found that when the ReconNet is randomly initialized, jointly training the ReconNet and the Generator does not converge.
It is likely that there exist configurations where the joint training converges without pretraining the ReconNet.
However, since jointly training the ReconNet and the Generator is computationally expensive, pretraining the ReconNet should still be desirable for faster convergence.

\section{Further details on evaluation}
\label{app:eval_details}
\textbf{Self-reenactment.}
For self-reenactment, we choose the first frame of the video as the reference frame and transfer all of the 3DMM parameters extracted from the video to drive the reference frame.
When computing L1, PSNR, and SSIM, we mask out the background because the background region is not of our interest.
FID is computed using pytorch-fid~\cite{Seitzer2020FID}.
CSIM is computed by averaging all of the ArcFace embeddings for the original video and the generated video.

\noindent
\textbf{Cross-reenactment.}
For cross-reenactment, we select random pairs from the dataset and use the first video to drive the first frame of the second video.

\noindent
\textbf{Lip-sync.}
For lip-sync, we use the pipeline outlined in Fig.~\ref{fig:fig4} of the main text.
We would like to note that there is a lip-sync quality gap when the original audio is used for lip-sync and when different audio (audio from a different video in the dataset) is used for lip-sync, see LSE-D and LSE-C in Table~\ref{tab:lipsync}.
We hypothesize that the quality gap is caused by a mismatch between the style blendshapes and the talking style implied by the audio.
To understand this phenomenon, let us consider for simplicity just the speaking speed.
During training, we sample the style blendshape from a different part of the video, so that the speaking speed is usually similar.
However, during inference, the speaking speed between the style blendshapes and the dubbing audio can be quite different, which is probably infrequent in the training data.

\section{Limitation}
\textbf{Limitations of 2D warping.}
For the THS, we borrowed the architecture of` the HeadGAN framework, which relies on 2D feature warping.
Feature warping offers a simple way to achieve temporally consistent video generation because convolutional neural nets are equivariant under translation, and feature warping is locally a translation.
However, the 2D warping is weak to occlusions, such as changes in head pose, visibility of teeth, and visibility of iris.
This is because it is difficult to encode the information of the occluded area into 2D features.
We expect that such limitations can be handled by 3D warping-based methods.
Fortunately, this is not difficult to achieve: our preliminary experiments indicate that by binding 3D keypoints to vertices of the 3DMMs, one can incorporate methods based on 3D warping fields into our approach.

\noindent
\textbf{Limitations in BlendingNet.}
We found that our BlendingNet is weak to large changes in illumination.
This becomes noticeable when there is strong contrast in the video.
We expect that this problem can be addressed by using diffusion inpainting methods, especially video diffusion inpainting methods, which work better than GAN based methods~\cite{saharia2022palette,lugmayr2022repaint,zhang2024avid}.

\noindent
\textbf{Limitations of the learned 3DMM parameters.}
The 3DMM parameters learned in our work is not perfect.
For example, it is not 3D consistent because we have not utilized multi-view datasets.
The teeth information is also not correctly encoded into the blendshapes.
This can be seen by changing the \texttt{mouthFunnel} blendshape and the \texttt{mouthUpperUp} blendshape.
Ideally, the teeth location should be the same when either of the blendshapes is activated, but we find that the teeth location is slightly different in Fig.~\ref{fig:fig5}.
This problem should also be solvable within our framework by labeling parts of the dataset with teeth landmarks (the full ICT-FaceKit mesh contains teeth), and incorporating these data when computing the landmark loss.

\noindent
\textbf{Limitations of identity and expression disentanglement.}
Although cross-reenactment was not the focus of our work, directly transferring all of the blendshapes from the driving video has limitations in keeping the source frame's identity.
This is because it is impossible to completely disentangle expression from identity, especially when using a single image.
We expect additional designs for retargeting will be helpful for cross-reenactment ~\cite{wang2021one,guo2024liveportrait}.

\noindent
\textbf{Limitations in capturing detailed expressions.}
Although some of the detailed expressions, such as wrinkles, can be captured with just the 3DMM (for example, the first row of the self-reenactment results of Fig.~\ref{fig:fig6} in main text), there are limitations.
It would be interesting to see if it is possible to supplement the 3DMM with additional features responsible for detailed changes to the face.

\noindent
\textbf{Limitations on Audio-to-Blendshape model.}
As mentioned in Sec.~\ref{app:eval_details}, the Audio-to-Blendshape model struggles when there is a mismatch between the style blendshapes and the audio.
To solve this problem, a more sophisticated treatment of style seems necessary.
% 325
\end{document}